# Longitudinal and Graph-Augmented Prediction of Adolescent Substance Use Onset in the ABCD Study

Yixuan He[1] Jinni Su[1] Yun Kang[1]
[1] Arizona State University
Yixuan.He@asu.edu, jinnisu1@asu.edu, Yun.Kang@asu.edu

**Abstract**

Early identification of adolescent substance-use risk is an important prevention challenge, yet the relative value of baseline characteristics, longitudinal trajectories, and relational context remains unclear. Using data from approximately 11,860 participants in the Adolescent Brain Cognitive Development (ABCD) Study, we compare cross-sectional, longitudinal, and graph-based approaches for predicting alcohol sipping, alcohol use, marijuana use, and alcohol/marijuana use. We evaluate tree-based models, recurrent neural networks, and Temporal Graph Convolutional Networks (T-GCNs) constructed from family, school, and feature-similarity graphs. Longitudinal models consistently outperform baseline models, with temporal XGBoost achieving the strongest standalone performance. Although T-GCNs generally do not surpass temporal XGBoost, graph-derived risk scores provide complementary information. Combining temporal XGBoost and T-GCN predictions through score-level stacking yields the best performance across all outcomes, achieving AUC-ROC values above 0.79. Feature analyses identify peer deviance, age, externalizing symptoms, parental monitoring, cultural norms, and neighborhood context as important predictors of substance use onset. These findings demonstrate the value of longitudinal modeling for substance-use prediction and suggest that graph-based representations can provide effective auxiliary risk signals.

## 1 Introduction

Adolescent substance-use initiation is a major public-health concern because early onset of substance use is associated with elevated risk of later substance use disorders and adverse developmental outcomes (C.-Y. Chen et al., 2009; Donovan & Molina, 2011; Poudel & Gautam, 2017). Delaying initiation substantially reduces subsequent risk, making early identification of high-risk youth an important prevention objective. The Adolescent Brain Cognitive Development (ABCD) Study provides a unique opportunity to study this problem using repeated assessments spanning individual, family, peer, school, neighborhood, and cultural domains (Volkow et al., 2018). Because these risk and protective factors evolve throughout adolescence, substance-use vulnerability should be viewed as a dynamic developmental process rather than a static baseline characteristic.

Recent machine-learning studies have successfully identified predictors of substance use and related developmental outcomes in ABCD. However, two important limitations remain. First, most approaches rely primarily on baseline or cross-sectional representations and therefore cannot fully exploit longitudinal trajectories (Niklason et al., 2025; Wei et al., 2026). Second, existing models generally treat participants as independent observations, despite extensive evidence that peer and social environments influence adolescent behavior (Hoang, 2026).

To address these limitations, we evaluate both temporal and graph-based approaches for predicting substance-use onset. We compare baseline machine-learning models, longitudinal sequence models, and Graph Convolutional Networks (GNNs) (Kipf & Welling, 2017; Zhao et al., 2019) across four outcomes: alcohol sipping, alcohol use, marijuana use, and alcohol/marijuana use. We further investigate whether graph-derived risk scores provide complementary information beyond strong temporal tabular learners. Using approximately 11,860 participants across up to ten temporal waves (time steps), we show that longitudinal modeling substantially improves prediction and that graph-based risk scores further improve performance when combined with temporal XGBoost through score-level stacking. Our main contributions are threefold: (1) a systematic comparison of baseline and longitudinal prediction frameworks for substance-use onset, (2) multiple graph-construction strategies capturing family, school, and feature-similarity relationships, and (3) a graph-augmented stacking framework that achieves the strongest predictive performance AUC-ROC (>0.79) across all outcomes.

# 2 Related Work

Machine-learning approaches have been applied to the ABCD study to identify predictors of substance use. Niklason et al. (2025) used XGBoost (T. Chen & Guestrin, 2016) and SHAP (Lundberg & Lee, 2017) to predict early alcohol sipping from baseline data, identifying family norms, socioeconomic factors, and sleep-related variables as key predictors. Wei et al. (2026) analyzed longitudinal ABCD data and demonstrated that time-varying environmental and behavioral factors contribute to substance-use initiation. Hoang (2026) used Elastic Net (Zou & Hastie, 2005) and Random Forest (Breiman, 2001) to predict internalizing disorder onset. Related studies have also successfully modeled conduct problems (Berluti et al., 2026), delay discounting trajectories (Kahn et al., 2026), and substance-use vulnerability (Moreira et al., 2026), highlighting the value of longitudinal prediction. However, these approaches generally do not explicitly incorporate relational structure. Graph neural networks offer a principled framework for integrating contextual information. Graph Convolutional Networks (GCNs) (Kipf & Welling, 2017) learn node representations by aggregating neighborhood (either through known family/school/site relationship or from inferred feature proximity) information, while Temporal Graph Convolutional Networks (T-GCNs) (Zhao et al., 2019) combine graph convolutions with recurrent units such as GRUs (Cho et al., 2014) to jointly model relational and temporal dependencies. Although graph-based methods have been successfully applied in neuroscience (Wallace & Conner, 2025), their use for adolescent substance-use prediction remains limited. Our work bridges this gap by systematically comparing baseline and longitudinal models and evaluating whether graphs provide complementary predictive value beyond strong temporal tabular learners.

# 3 Data Description, Preparation, and Prediction Tasks

The ABCD Study is a large, multi-site longitudinal cohort following children through adolescence (Garavan et al., 2018). We analyzed approximately 11,860 participants with up to T=10 annual or mid-year waves (time steps), with mean age ranging from 10 years (SD = .62) at T1 to 17 years (SD = .66) at T10. The modeling table included 48 standardized continuous or ordinal Z features spanning sociodemographic (e.g., age, parent education, family income), individual (e.g., cognitive functioning, mental health), family (e.g., parenting, family relationships), peer (e.g., affiliation with deviant or prosocial peers), school (e.g., school environment, academic engagement), neighborhood (e.g., deprivation, alcohol outlet density), cultural (e.g., cultural values, ethnic identity) and structural (e.g., state law on marijuana use, state-level immigration bias) domains. Missing values (~55% of feature-wave cells) were handled via Full-Information Maximum Likelihood with EM (FIML-EM) (Enders, 2022). Missingness primarily reflects unequal measure availability across waves and assessments rather than participant attrition alone, although non-random missingness remains a limitation. We also consider categorical demographic indicators including sex and race. The feature dimension for training and evaluation is F=53. Stable ID fields used for graph construction, including family and school identifiers, are forward filled within participant when later waves were missing an ID value, without overwriting nonempty later values. Outcomes were final-wave onset of alcohol sipping, alcohol use (i.e., having consumed a full drink of alcohol), marijuana use, and alcohol/marijuana use (defined as being positive if any of alcohol sipping, alcohol use, or marijuana use variable is positive). Approximate final-wave prevalence in the analysis set is 55% for alcohol sipping, 35% for alcohol use, 28% for marijuana use, and 60% for alcohol/marijuana use.

We evaluate two prediction tasks. Task 1 is baseline-only: baseline wave-1 features are used to predict the final-wave outcome. Task 2 is temporal: all waves before the final outcome wave are used to predict the final-wave outcome. For every outcome and random seed, participants were split into stratified 70% training, 15% validation, and 15% test sets using seeds 0, 10, 20, 30, and 40.

To assess whether graph structures may be beneficial, we construct 17 graph variants. The graph families are chosen to represent distinct developmental mechanisms. Family graphs capture shared genetic and household environmental influences. School graphs approximate peer exposure and shared institutional environments. Site graphs capture regional and recruitment-related contextual effects. Feature-similarity graphs (kNN and cosine) represent latent homophily, connecting participants with similar behavioral, demographic, and environmental profiles even when explicit social relationships are unavailable. Multiple sparsity levels (k = 5, 10, 20 and cosine thresholds of 0.5%, 1%, 5%, and 10%) are evaluated to examine the sensitivity of graph aggregation to neighborhood size. Directed kNN preserves asymmetric nearest-neighbor relations, whereas symmetric kNN approximates mutual or bidirectional similarity relationships.

Specifically, we construct seven ID graphs connecting participants sharing school, site, family, or the union of two or three of these IDs: school, site, family, school+site, school+family, site+family, and school+site+family. Family graphs are kept sparse by excluding implausibly large family groups, whereas school and site groups are allowed to be large by design; three directed k-nearest-neighbor graphs, kNN5, kNN10, and kNN20, connect each participant to its nearest neighbors under a missingness-aware Euclidean distance; three symmetric kNN graphs, kNNsym5, kNNsym10, and kNNsym20, first select directed neighbors and then symmetrize the edge support (for kNNsym5, an edge is constructed if the similarity level is among the top five for either endpoint); and four sparse cosine graphs retain the top 0.5%, 1%, 5%, or 10% of masked cosine-similarity pairs. All graphs include self-loops, are binarized to contain only 0/1 edge weights, and have row-normalized adjacency matrices. All graphs are constructed separately for each wave, with feature graphs constructed using standardized features together with categorical demographic features under missingness-aware distance calculations. For feature-similarity graphs, to ensure fair similarity computation, we expand the one-hot encodings for sex and race to include the linearly dependent dimension as well (which is removed during model training and evaluation to avoid multicollinearity), resulting in a graph-construction feature dimension of F+2=55.

# 4 Experiments

## 4.1 Methods and Experimental Setup

We evaluate different machine learning approaches on our two prediction tasks on four different outcomes. For Task 1, we include baseline methods Elastic Net, Random Forest, XGBoost, and MLP (256-128-1, Dropout 0.2). Our proposed graph neural network approaches are based on different graphs applied to a GCN model (2 layers, 64-dim hidden, sparse adjacency). Task 2 baselines include LSTM (Hochreiter & Schmidhuber, 1997), GRU, XGBoost-temporal (flattened T×F input). Temporal XGBoost flattened the longitudinal feature tensor and fit a gradient-boosted tree classifier, allowing nonlinear thresholds and interactions across waves without imposing a recurrent state structure. Task 2 graph models used T-GCN graph convolution (2-layer GCN spatial encoder) at each wave followed by a GRU temporal update (64-dim). Unlike the original T-GCN application with a fixed traffic network, we construct a graph at each wave. The graph convolution therefore uses the wave-specific adjacency matrix, while the GRU models temporal evolution of participant representations across waves. For graph-augmented stacking, each trained T-GCN model produces one risk score for each individual per graph. The default stacker uses four features: temporal XGBoost risk, T-GCN-family risk, T-GCN-kNNsym5 risk, and T-GCN-cos0.5% risk. Family graphs generally perform best among explicit relationship graphs, while graph smoothing can reduce performance in denser feature-similarity graphs. We therefore select the strongest family graph together with relatively sparse representatives from the kNN and cosine graph families. We evaluate a logistic score stacker (applying a logistic regression model to these four risk scores, denoted *XGB+TGCN score-logit*), an XGBoost score stacker (applying XGBoost to these four risk scores, denoted *XGB+TGCN score-XGB*), and an exploratory augmented XGBoost model that appends the three T-GCN risk scores to the original flattened Task 2 features and further applies a full XGBoost model to these augmented input features (denoted *XGB+TGCN augmented*). For all neural models, we employ the Adam optimizer (lr=1e-3, weight decay=1e-4), with a maximum of 1,000 epochs, following an early stopping scheme (patience=200 epochs without validation-loss improvement) with best-weight restoration. We use Binary Cross-Entropy loss with class imbalance weighting.

To support domain interpretation, we summarize feature importance. XGBoost and temporal XGBoost use mean absolute SHAP values, Random Forest uses impurity-based importance, Elastic Net uses coefficient magnitude, and GNN models use population-level gradient x input saliency average across test participants and, for T-GCN, across time. Domain importance is computed by grouping feature names into demographic, individual, family, peer, school, neighborhood, cultural, and structural domains and averaging importance values within each domain.

FIML-EM is performed before train/validation/test splitting using predictor variables only and does not use outcome labels. Likewise, feature-similarity graphs are constructed from predictor features only. Our GNN experiments follow the standard transductive setting, where all nodes are present during graph construction and message passing, but labels from validation and test participants are masked and never used for optimization. Consequently, information from test participants may influence graph structure through unlabeled feature similarity, but no outcome information is shared across splits.

Experiments were conducted on one compute node with two Nvidia A100 GPUs with driver version 595.71.05 and CUDA version 13.2, 48 AMD EPYC 74F3 24-Core Processor CPUs and 503GB RAM. Anonymized code is provided in https://anonymous.4open.science/status/ABCD-substance. Results are averaged over five random seeds, with one standard deviation reported as well.

## 4.2 Results

We evaluate the machine learning methods using class-imbalance-aware metrics including AUC-ROC (both tasks) and AUC-PR (Task 2, precision-recall curve). Because marijuana use prevalence is approximately 28%, AUC-PR provides a complementary assessment under class imbalance. We further provide feature and domain importance analysis for interpretation and insights.

Table 1 summarizes Task 1 AUC-ROC performance with selected GCN methods (full set report omitted). Standard machine learning models like XGBoost and Random Forest consistently outperformed baseline Graph Neural Networks (GNNs) and Multilayer Perceptrons (MLP). For example, in predicting "Alcohol Sips Ever," XGBoost achieved an AUC-ROC of 0.706, compared to GCN models which hovered around 0.62–0.69. The strongest cross-sectional models were the tree baselines, especially XGBoost and Random Forest. GCN variants are competitive in some outcomes but do not consistently exceed the non-graph baselines, suggesting that graph smoothing at baseline does not add enough information beyond individual-level baseline features.

| Model | Alcohol sips | Alcohol | Marijuana | Alcohol/ Marijuana |
|---|---|---|---|---|
| Elastic Net | 0.684±0.008 | 0.680±0.014 | 0.638±0.005 | 0.669±0.011 |
| Random Forest | 0.703±0.007 | **0.691±0.015** | **0.643±0.007** | **0.695±0.013** |
| XGBoost | **0.706±0.008** | 0.691±0.016 | 0.642±0.005 | 0.694±0.014 |
| MLP | 0.689±0.013 | 0.672±0.015 | 0.623±0.004 | 0.677±0.015 |
| GCN-family | 0.694±0.012 | 0.658±0.067 | 0.608±0.065 | 0.684±0.017 |
| GCN-school | 0.680±0.015 | 0.671±0.011 | 0.591±0.072 | 0.675±0.019 |
| GCN-kNNsym5 | 0.683±0.016 | 0.632±0.049 | 0.591±0.050 | 0.672±0.019 |
| GCN-cos0.5% | 0.680±0.012 | 0.631±0.037 | 0.574±0.041 | 0.662±0.013 |

**Table 1: Task 1 AUC-ROC** performance on baseline to final onset prediction. Results are averaged over five random seeds in terms of splits and training, plus/minus one standard deviation. The best is marked in **bold**.

Table 2 and Table 3 summarize Task 2 performance. Predictive performance increases substantially when incorporating temporal data. GRU, LSTM, and temporal XGBoost achieve AUC-ROC values between 0.78 and 0.82 across outcomes. Across outcomes, temporal XGBoost consistently improves over baseline XGBoost, with AUC-ROC gains ranging from 0.086 to 0.171. The score-logit XGBoost+TGCN stacker achieves the highest AUC-ROC for all four Task 2 outcomes.

Because several outcomes are moderately imbalanced (e.g., marijuana-use prevalence ≈28%), we additionally report AUC-PR in Table 3. The ranking of methods is largely consistent with the AUC-ROC results. The score-logit XGBoost+TGCN stacker also achieves the highest AUC-PR for all four Task 2 outcomes. For marijuana use, temporal XGBoost achieves an AUC-PR of 0.634, while the score-logit stacker improves this to 0.646. For the composite alcohol/marijuana outcome, AUC-PR increases from 0.856 to 0.860. These findings suggest that the gains observed in AUC-ROC are not solely attributable to class imbalance.

Together, these results indicate that developmental trajectories contain information that cannot be fully recovered from a baseline snapshot alone. The gains from the score-logit stacker are modest but consistent, suggesting that T-GCNs and graph structures capture a small amount of complementary risk information. Directly appending T-GCN scores to the full flattened feature set often underperforms score-level stacking. This may occur because the raw temporal feature space already provides XGBoost with many high-resolution predictors, whereas score-level stacking operates as a calibrated ensemble over distinct risk summaries. In practice, graph scores may therefore be more useful as late-fusion predictors than as additional raw columns in an already high-dimensional boosted tree.

Across graph connectivity patterns, family and school graphs generally outperform site-based graphs. The relatively poor performance of the site graph suggests that immediate social environments and participant similarity are more informative than broad regional location.

| Model | Alcohol sips | Alcohol | Marijuana | Alcohol/ Marijuana |
|---|---|---|---|---|
| LSTM | 0.781±0.007 | 0.800±0.007 | 0.795±0.016 | 0.784±0.016 |
| GRU | 0.788±0.007 | 0.808±0.010 | 0.808±0.012 | 0.793±0.015 |
| Temporal XGBoost | 0.792±0.011 | 0.819±0.007 | 0.813±0.013 | 0.797±0.016 |
| T-GCN school | 0.742±0.007 | 0.752±0.008 | 0.735±0.007 | 0.742±0.010 |
| T-GCN site | 0.644±0.016 | 0.639±0.018 | 0.533±0.053 | 0.649±0.018 |
| T-GCN family | 0.776±0.006 | 0.792±0.008 | 0.792±0.016 | 0.779±0.016 |
| T-GCN school+site | 0.651±0.016 | 0.655±0.020 | 0.595±0.014 | 0.654±0.019 |
| T-GCN school+family | 0.740±0.006 | 0.748±0.006 | 0.735±0.006 | 0.740±0.011 |
| T-GCN site+family | 0.644±0.015 | 0.639±0.020 | 0.533±0.053 | 0.648±0.020 |
| T-GCN school+site+family | 0.653±0.017 | 0.654±0.021 | 0.596±0.018 | 0.654±0.022 |
| T-GCN kNN5 | 0.720±0.018 | 0.729±0.008 | 0.700±0.006 | 0.716±0.017 |
| T-GCN kNN10 | 0.713±0.018 | 0.713±0.012 | 0.675±0.014 | 0.711±0.022 |
| T-GCN kNN20 | 0.710±0.018 | 0.704±0.010 | 0.659±0.011 | 0.707±0.021 |
| T-GCN kNNsym5 | 0.757±0.011 | 0.770±0.013 | 0.759±0.007 | 0.757±0.017 |
| T-GCN kNNsym10 | 0.753±0.012 | 0.757±0.014 | 0.748±0.007 | 0.757±0.020 |
| T-GCN kNNsym20 | 0.741±0.016 | 0.742±0.014 | 0.727±0.008 | 0.737±0.024 |
| T-GCN cos0.5% | 0.716±0.008 | 0.714±0.015 | 0.691±0.016 | 0.707±0.017 |
| T-GCN cos1% | 0.712±0.010 | 0.713±0.011 | 0.687±0.017 | 0.702±0.021 |
| T-GCN cos5% | 0.727±0.007 | 0.732±0.011 | 0.717±0.009 | 0.738±0.021 |
| T-GCN cos10% | 0.744±0.007 | 0.754±0.011 | 0.744±0.009 | 0.766±0.014 |
| XGB+TGCN score-logit | **0.796±0.010** | **0.822±0.007** | **0.819±0.014** | **0.802±0.017** |
| XGB+TGCN score-XGB | 0.794±0.009 | 0.820±0.008 | 0.815±0.014 | 0.801±0.017 |
| XGB+TGCN augmented | 0.786±0.009 | 0.804±0.009 | 0.800±0.017 | 0.790±0.016 |

**Table 2: Task 2 AUC-ROC** performance on sequential to final onset prediction. Results are averaged over five random seeds in terms of splits and training, plus/minus one standard deviation. The best is marked in **bold**.

| Model | Alcohol sips | Alcohol | Marijuana | Alcohol/Marijuana |
|---|---|---|---|---|
| LSTM | 0.813±0.006 | 0.661±0.017 | 0.617±0.026 | 0.845±0.012 |
| GRU | 0.820±0.006 | 0.673±0.019 | 0.633±0.022 | 0.853±0.012 |
| Temporal XGBoost | 0.821±0.010 | 0.687±0.017 | 0.634±0.025 | 0.856±0.014 |
| T-GCN family | 0.808±0.004 | 0.651±0.016 | 0.611±0.035 | 0.841±0.013 |
| T-GCN school | 0.767±0.008 | 0.585±0.029 | 0.515±0.026 | 0.804±0.012 |
| T-GCN kNNsym5 | 0.789±0.010 | 0.614±0.017 | 0.567±0.022 | 0.824±0.014 |
| T-GCN cos0.5% | 0.753±0.005 | 0.553±0.024 | 0.470±0.021 | 0.768±0.009 |
| XGB+TGCN score-logit | **0.826±0.009** | **0.697±0.017** | **0.646±0.030** | **0.860±0.015** |
| XGB+TGCN score-XGB | 0.821±0.007 | 0.689±0.021 | 0.638±0.034 | 0.858±0.013 |
| XGB+TGCN augmented | 0.818±0.006 | 0.670±0.022 | 0.621±0.034 | 0.851±0.015 |

**Table 3: Task 2 AUC-PR** performance on sequential to final onset prediction. Results are averaged over five random seeds in terms of splits and training, plus/minus one standard deviation. The best is marked in **bold**.

Table 4 summarizes the most interpretable importance results. Across Task 1 XGBoost models, age, family or cultural rules around substance use, structural context, and neighborhood features are repeatedly selected and emphasized. In Task 2, the strongest temporal XGBoost predictors concentrate at later waves, especially deviant peer exposure, age, and externalizing problems. This pattern supports the performance results: longitudinal behavioral and peer-context trajectories contain risk information that a baseline-only model cannot fully recover.

The graph-model importance summaries (omitted here) are broadly consistent with the tabular models. Family and kNNsym5 T-GCNs most often emphasize age, externalizing problems, peer deviance, alcohol rules, and family recreation or supervision features. In the score-level stacker, the temporal XGBoost score remains the largest contributor, but family and kNNsym5 T-GCN risk scores are consistently nonzero and usually larger than the sparse cosine graph score. This supports a domain interpretation in which graph models add small relational/contextual risk summaries on top of the dominant individual trajectory signal.

| Outcome | Model | Leading domains | Leading features or score terms |
|---|---:|---:|---:|
| Alcohol sips | Task 1 XGBoost | Demographic (Z) 0.136; Structural 0.100; Neighborhood 0.085 | Cultural Religion AlcoholRules 0.237; Demographic ChildAge 0.190; State ImmigrationBias 0.178 |
| Alcohol | Task 1 XGBoost | Demographic (Z) 0.178; Neighborhood 0.101; Cultural 0.080 | Demographic ChildAge 0.331; Cultural Religion AlcoholRules 0.212; Demographic NonHispanic Black 0.160 |
| Marijuana | Task 1 XGBoost | Demographic (Z) 0.155; Neighborhood 0.096; Family 0.075 | Demographic ChildAge 0.254; Cultural value Religion p 0.180; Demographic ParentEducation 0.108 |
| Alcohol/ Marijuana | Task 1 XGBoost | Demographic (Z) 0.140; Structural 0.107; Neighborhood 0.085 | Cultural Religion AlcoholRules 0.236; Demographic ChildAge 0.230; State ImmigrationBias 0.179 |
| Alcohol sips | Task 2 temporal XGBoost | Other 0.030; Demographic (Z) 0.024; Individual 0.017 | Peer friends deviant (t8) 0.218; Individual ExternalizingProblems (t8) 0.201; Demographic ChildAge (t8) 0.181 |
| Alcohol | Task 2 temporal XGBoost | Other 0.044; Demographic (Z) 0.033; Individual 0.019 | Peer friends deviant (t8) 0.376; Demographic ChildAge (t8) 0.329; Demographic ChildAge (t6) 0.184 |
| Marijuana | Task 2 temporal XGBoost | Other 0.064; Demographic (Z) 0.027; Peer 0.024 | Peer friends deviant (t8) 0.483; Demographic ChildAge (t8) 0.303; Individual ExternalizingProblems (t7) 0.220 |
| Alcohol/ Marijuana | Task 2 temporal XGBoost | Other 0.036; Demographic (Z) 0.025; Individual 0.016 | Peer friends deviant (t8) 0.262; Demographic ChildAge (t8) 0.239; Individual ExternalizingProblems (t7) 0.155 |

**Table 4: Feature and domain importance** for selected interpretable models.

Investigating into importance analysis for all methods (full results omitted), the models identify a consistent set of high-impact features across substance types: (1) Demographic Factors: Child Age was the most dominant predictor in nearly every model and task. (2) Cultural and Religious Factors:

Religious rules regarding alcohol (ZCultural_Religion_AlcoholRules) and the general value placed on religion were top predictors, particularly for alcohol-related outcomes. (3) Environmental Factors: Neighborhood deprivation and structural features like immigration bias were consistently ranked in the top 10 for cross-sectional models. (4) Peer and Behavioral Shift (Longitudinal): In Task 2, new predictors became critical. Deviant peer affiliation (Zpeer_friends_deviant) and individual externalizing problems (behavioral issues) emerged as top-tier predictors alongside age. Further, the models categorize features into domains to identify which "layers" of the adolescent system are most influential. For Task 1, the prediction is dominated by Demographic, Structural, Neighborhood, and Cultural domains. For Task 2, the leading influence shifts toward Individual and Other (which includes peer measures) as the adolescent ages and transitions through the study waves.

## 4.3 Discussions

Our results demonstrate that longitudinal information is the primary driver of predictive performance for adolescent substance-use onset. Across all outcomes, temporal models substantially outperformed baseline approaches, indicating that developmental trajectories contain important risk signals that cannot be fully recovered from a single baseline assessment. In particular, temporal XGBoost consistently achieved the strongest standalone performance, suggesting that flexible nonlinear modeling of longitudinal factors is highly effective for this prediction task.

Although graph-based models generally did not surpass temporal XGBoost, they capture complementary information that improves performance when incorporated through score-level stacking. The strongest graph variants are typically family and feature-similarity graphs, suggesting that relational context contributes additional risk information beyond individual trajectories.

Feature importance analyses revealed a consistent set of predictors across outcomes, including age, peer deviance, externalizing behaviors, cultural substance-use norms, and family-related factors.

Overall, our results suggest that future substance-use risk is best characterized through a combination of individual developmental trajectories and relational context. While temporal information remains the dominant source of predictive signal, graph-derived representations provide modest but consistent improvements and offer a promising direction for incorporating social and contextual structure into adolescent risk prediction. On the other hand, practical deployment would require selecting operating thresholds based on intervention costs, resource constraints, and acceptable false-positive rates, which is left as future work. The composite alcohol/marijuana outcome increases statistical power but combines behaviors with different developmental significance and risk severity.

# 5 Conclusion and Future Work

In ABCD substance-use prediction, longitudinal trajectories are the dominant source of predictive improvement: temporal XGBoost substantially boosts baseline XGBoost. Graph neural networks are informative but usually weaker than temporal XGBoost, especially when graph aggregation dilutes individual trajectories. In our empirical studies, graph-specific T-GCN risk scores can modestly improve temporal XGBoost while preserving a clear interpretation of graph contribution and domain signal, achieving AUC-ROC of 0. 792–0.822 for predicting adolescent substance use onset across all four ABCD outcomes (i.e., onset of alcohol sips, alcohol use, marijuana use, and alcohol/marijuana use), substantially outperforming cross-sectional baselines (+0.086–+0.176) and recurrent-only models (+0.008–+0.024). Future work may include more complex graph neural network models incorporating the tree-based models' advantages, an incorporation of domain knowledge, and consideration of trajectory analysis.